\documentclass{article}

\usepackage{arxiv}

\usepackage[utf8]{inputenc} % allow utf-8 input
\usepackage[T1]{fontenc}    % use 8-bit T1 fonts
\usepackage{hyperref}       % hyperlinks
\usepackage{url}            % simple URL typesetting
\usepackage{booktabs}       % professional-quality tables
\usepackage{amsfonts,amsmath,amssymb}       % blackboard math symbols
\usepackage{nicefrac}       % compact symbols for 1/2, etc.
\usepackage{microtype}      % microtypography
\usepackage{lipsum}
\usepackage{algorithm}
\usepackage{algorithmic}
\usepackage{graphicx}
\usepackage{geometry}
\graphicspath{ {./figs/} }

\title{OverNaN: NaN-Aware Oversampling for Imbalanced Learning with Meaningful Missingness}

\author{
 Amanda S Barnard \\
  School of Computing\\
  Australian National University\\
  Acton, ACT 2601 \\
  \texttt{amanda.s.barnard@anu.edu.au} \\
}

\begin{document}
\maketitle
\begin{abstract}
Missing values are routinely treated as defects to be eliminated through deletion or imputation prior to machine learning. In many applied domains, however, missingness itself carries information, reflecting experimental constraints, measurement choices, or systematic mechanisms tied to the data-generating process. Eliminating or masking this structure can distort class boundaries, introduce bias, and reduce generalisability; particularly in imbalanced datasets where minority classes are already under-represented.  OverNaN is a lightweight, NaN-aware oversampling framework designed to address class imbalance without erasing missingness structure. It extends common synthetic oversampling methods to operate directly on incomplete feature vectors, allowing missing values to be preserved, propagated, or selectively interpolated according to explicitly defined strategies. Rather than repairing missing data, OverNaN treats missingness as part of the feature space over which synthetic samples are generated.  This paper situates OverNaN within the broader landscape of imbalanced learning, missing-data handling, and NaN-tolerant algorithms. Using representative examples included with the software, we demonstrate that meaningful missingness can be retained during oversampling without introducing artificial certainty. OverNaN is intended for practitioners working with small, incomplete, and imbalanced datasets in scientific and engineering domains where missingness is unavoidable and often informative.
\end{abstract}

% keywords can be removed

\section{Introduction}

Missing values are pervasive in applied machine learning, particularly in the physical sciences, engineering, biomedicine, and health. In many such domains, missingness arises from structural constraints rather than stochastic noise: measurements may be destructive, prohibitively expensive, ethically restricted, or conditional on earlier outcomes. As a result, the absence of data often reflects domain-relevant mechanisms rather than random failure.

Despite this, standard machine learning pipelines treat missing values as defects to be eliminated prior to analysis. Listwise deletion removes incomplete samples entirely, while imputation replaces missing entries with inferred values derived from statistical heuristics or predictive models \cite{little1987statistical,rubin1976inference}. Both approaches can distort downstream learning. Deletion disproportionately removes minority-class samples, exacerbating class imbalance, while imputation collapses uncertainty into point estimates, introducing artificial certainty and potentially embedding model-dependent bias.

These effects are particularly damaging in imbalanced learning, where the minority class is already under-represented. Synthetic oversampling techniques such as SMOTE \cite{chawla2002smote} and ADASYN \cite{he2008adasyn} have been widely adopted to mitigate imbalance, but they implicitly assume complete feature vectors. In practice, missing values are commonly imputed before oversampling, erasing missingness structure before synthetic data are generated.

At the same time, many modern learning algorithms (particularly tree-based ensemble methods) can natively accommodate missing values by learning explicit split directions for absent features \cite{chen2016xgboost,ke2017lightgbm}. This ability is typically exploited as an implementation convenience rather than a modelling choice. Missingness is tolerated, but rarely treated as informative.

The central premise of this work is that missingness itself may carry information, and that eliminating or masking it prior to resampling can result in loss of signal or accumulation of bias. OverNaN was developed to address this gap. Rather than repairing incomplete data, OverNaN performs synthetic oversampling directly in the presence of NaNs, allowing missingness patterns to be preserved, propagated, or selectively interpolated under explicit control.

The goal is not to replace imputation or missingness modelling, but to provide an alternative for settings in which missing values are structural, unavoidable, and potentially meaningful.

\section{Problem Setting and Background}

Let $\mathcal{D}=\{(\mathbf{x}_i, y_i)\}_{i=1}^N$ denote a labeled dataset, where
\[
\mathbf{x}_i \in (\mathbb{R} \cup \{\text{NaN}\})^d, \qquad y_i \in \{1,\dots,C\}.
\]
We consider the imbalanced classification setting, where at least one class $c^\ast$ satisfies
\[
|\{i : y_i = c^\ast\}| \ll |\{i : y_i = c\}| \quad \text{for some } c \neq c^\ast.
\]

Conventional oversampling methods generate synthetic points
\[
\tilde{\mathbf{x}} = \mathbf{x}_i + \lambda (\mathbf{x}_j - \mathbf{x}_i),
\quad \lambda \sim \mathcal{U}(0,1),
\]
for pairs of minority-class samples $\mathbf{x}_i, \mathbf{x}_j$, implicitly assuming all feature-wise operations are defined. When NaNs are present, either samples are discarded, or missing values are imputed prior to oversampling.

From a statistical perspective, this corresponds to conditioning synthetic sample generation on a completed dataset, rather than the observed data-generating process. If missingness is not random, this substitution alters the geometry of feature space and the effective decision boundary.

OverNaN retains NaNs during oversampling. The feature space is treated as partially defined, and operations are restricted to observed components. Missingness patterns are explicitly propagated rather than implicitly repaired.

\section{Background: Synthetic Oversampling Methods}

Synthetic oversampling methods address class imbalance by generating additional samples for under-represented classes. Rather than duplicating existing observations, these methods populate sparse regions of feature space using interpolation or stochastic sampling. The central assumption is that minority-class structure can be approximated locally using observed data.

Importantly, most oversampling algorithms were developed under the assumption of complete feature vectors. Missing values are therefore typically removed or imputed prior to application, even though this step is often treated as incidental rather than consequential.

\subsection{SMOTE}

The Synthetic Minority Over-sampling Technique (SMOTE) generates new minority-class samples by linear interpolation between existing observations \cite{chawla2002smote}. Let $\mathbf{x}_i$ be a minority-class sample, and let $\mathbf{x}_j$ be one of its $k$ nearest neighbours in feature space. A synthetic sample is generated as
\[
\tilde{\mathbf{x}} = \mathbf{x}_i + \lambda (\mathbf{x}_j - \mathbf{x}_i),
\quad \lambda \sim \mathcal{U}(0,1).
\]

This procedure is repeated for multiple neighbours and interpolation coefficients to populate minority regions without replicating exact samples. SMOTE is simple, effective, and widely adopted across domains where data are scarce and imbalance is moderate.

However, SMOTE implicitly assumes that distances and convex combinations are well defined in all feature dimensions. When missing values are present, practitioners typically apply imputation prior to oversampling. As a result, synthetic samples reflect interpolations between observed and inferred values, even when the latter were not directly measured. This can introduce artificial smoothness and distort class boundaries, particularly in small datasets.

\subsection{ADASYN}

Adaptive Synthetic Sampling (ADASYN) extends SMOTE by focusing sampling effort on minority-class samples that are more difficult to learn \cite{he2008adasyn}. Difficulty is quantified by the local density of majority-class neighbours. For each minority sample $\mathbf{x}_i$, the proportion of majority-class samples among its $k$ nearest neighbours is computed, yielding a difficulty score $r_i$.

Synthetic samples are then generated in proportion to $r_i$, such that regions with greater class overlap receive more synthetic points. Interpolation proceeds in the same manner as SMOTE, but the number of samples generated per instance is adaptively controlled.

ADASYN can improve decision boundary resolution in highly overlapping or heterogeneous feature spaces. However, it inherits the same implicit requirements as SMOTE. Neighbour relationships and interpolation rely on complete feature vectors, and missing values are normally addressed through preprocessing. In the presence of imputed data, the adaptive weighting may amplify uncertainty or bias introduced during missing data repair.

\subsection{ROSE}

Random Over-Sampling Examples (ROSE) takes a different approach, generating synthetic samples from a smoothed, class-conditional kernel density estimate \cite{menardi2014rose}. Rather than interpolating directly between nearest neighbours, ROSE perturbs observed samples using Gaussian noise, with bandwidth parameters controlling the extent of smoothing.

From a conceptual perspective, ROSE assumes that minority-class structure can be approximated by a continuous density rather than discrete point-to-point interpolations. This can be advantageous in high-dimensional or irregular feature spaces, where nearest-neighbour methods may be unstable.

As with SMOTE and ADASYN, ROSE is generally formulated for complete data. Kernel density estimation with missing values is non-trivial, and practical implementations typically rely on imputation or restriction to complete cases. When imputation is used, the resulting density reflects both observed measurements and inferred values, again collapsing uncertainty prior to sample generation.

\subsection{Conceptual Comparison}

The essential differences between SMOTE, ADASYN, ROSE, and OverNaN can be summarized by examining how each method defines neighbourhoods, generates synthetic samples, and treats missing values.

SMOTE operates on explicit pairwise interpolation. Given minority samples $\mathbf{x}_i$ and $\mathbf{x}_j$, synthetic points are restricted to the line segment connecting them, $\mathbf{x}_i + \lambda(\mathbf{x}_j - \mathbf{x}_i)$, with $\lambda \in (0,1)$. The geometry is strictly local and linear, and interpolation implicitly assumes all feature dimensions are observed. Missing values are therefore incompatible with the method unless repaired beforehand.

ADASYN retains the same interpolation rule as SMOTE, but modifies the sampling distribution. The expected number of synthetic samples generated around $\mathbf{x}_i$ is proportional to a density-based difficulty measure, defined by the proportion of neighbouring majority-class instances. Formally, interpolation remains linear in feature space, but the probability measure over interpolation sites is adaptive rather than uniform. As with SMOTE, distances and densities are only well defined for complete feature vectors.

ROSE replaces explicit interpolation with sampling from a smoothed class-conditional distribution. Synthetic samples are drawn from
\[
\tilde{\mathbf{x}} \sim \mathcal{N}(\mathbf{x}_i, \Sigma),
\]
where $\Sigma$ is typically diagonal or isotropic and scaled by a bandwidth parameter. The resulting geometry is distributional rather than pairwise, and synthetic points are not constrained to convex combinations of observed instances. Kernel density estimation, however, presumes fully observed feature vectors, and practical use again relies on prior imputation or deletion.

SMOTE, ADASYN, and ROSE represent complementary strategies for generating synthetic minority-class samples. All three are effective under appropriate conditions and have been validated extensively in applied settings. This design choice is rarely questioned, but it has important consequences. In datasets where missingness is systematic or informative, imputation or deletion performed prior to oversampling can alter feature distributions, distort local geometry, and introduce assumptions that are subsequently amplified during synthetic sample generation.

OverNaN modifies none of the core geometric assumptions of these methods. Neighbour-based and kernel-based formulations are retained where applicable. The key distinction lies in the domain of definition. Feature space is treated as $(\mathbb{R} \cup \{\text{NaN}\})^d$, rather than $\mathbb{R}^d$. Distances are computed on shared observed subspaces, and synthetic samples are generated feature-wise, with missing values preserved or propagated according to explicit rules rather than resolved implicitly.  In this sense, OverNaN does not introduce a new sampling geometry. It alters the admissible operations within existing geometries by refusing to substitute inferred values where observations are absent.

\section{Design Principles}

OverNaN is guided by three principles:
\begin{enumerate}
\item First, missing values are treated as structural elements of the data, not defects. Synthetic samples may legitimately contain NaNs if the original data support such patterns.
\item Second, oversampling should not introduce artificial certainty. Generating complete synthetic samples from incomplete real data risks encoding assumptions that are neither observed nor justified.
\item Third, control over missingness should be explicit. Different applications warrant different treatments of NaNs, and these choices should be visible and deliberate rather than implicit side-effects of preprocessing.
\end{enumerate}

\subsection{NaN-Aware Oversampling}

OverNaN implements NaN-aware variants of common oversampling strategies. The key design principle is that synthetic samples should not contain more information than the data that generated them.

Let $\mathbf{x}_i, \mathbf{x}_j$ be two samples, and define the set of observed dimensions
\[
\Omega_{ij} = \{k \in \{1,\dots,d\} : x_{ik} \neq \text{NaN} \land x_{jk} \neq \text{NaN}\}.
\]

Distances are computed only on shared observed dimensions,
\[
\|\mathbf{x}_i - \mathbf{x}_j\|_{\Omega_{ij}} =
\left( \sum_{k \in \Omega_{ij}} (x_{ik}-x_{jk})^2 \right)^{1/2},
\]
ensuring neighbour relationships are well-defined without imputation.

Synthetic samples are generated feature-wise,
\[
\tilde{x}_k =
\begin{cases}
x_{ik} + \lambda (x_{jk} - x_{ik}), & k \in \Omega_{ij}, \\
\text{NaN or rule-based assignment}, & k \notin \Omega_{ij},
\end{cases}
\]
where the treatment of missing features depends on the chosen strategy.

\subsection{Missingness Handling Strategies}

OverNaN implements three explicit strategies.

\paragraph{Preserve-Pattern Strategy.}
A feature is marked as missing in the synthetic sample if it is missing in either parent. This is conservative and preserves observed missingness structure (see Algorithm~\ref{alg1}).

\begin{algorithm}[H]
\caption{NaN-Preserving Oversampling \label{alg1}}
\begin{algorithmic}
\STATE Select minority sample $\mathbf{x}_i$
\STATE Identify neighbour $\mathbf{x}_j$
\FOR{each feature $k$}
    \IF{$x_{ik} = \text{NaN}$ OR $x_{jk} = \text{NaN}$}
        \STATE $\tilde{x}_k \leftarrow \text{NaN}$
    \ELSE
        \STATE $\tilde{x}_k \leftarrow x_{ik} + \lambda (x_{jk}-x_{ik})$
    \ENDIF
\ENDFOR
\end{algorithmic}
\end{algorithm}

\paragraph{Selective Interpolation Strategy.}
Observed values are interpolated when available; missing values arise only when neither parent provides information (see Algorithm~\ref{alg2}).

\begin{algorithm}[H]
\caption{Selective NaN Interpolation \label{alg2}}
\begin{algorithmic}
\FOR{each feature $k$}
    \IF{$x_{ik} \neq \text{NaN}$ AND $x_{jk} \neq \text{NaN}$}
        \STATE $\tilde{x}_k \leftarrow x_{ik} + \lambda (x_{jk}-x_{ik})$
    \ELSE
        \STATE $\tilde{x}_k \leftarrow \text{NaN}$
    \ENDIF
\ENDFOR
\end{algorithmic}
\end{algorithm}

\paragraph{Probabilistic Missingness Strategy.}
Missingness is sampled according to empirical per-class missingness rates. Let $p_k^{(c)}$ be the probability that feature $k$ is missing in class $c$ (see Algorithm~\ref{alg3}).

\begin{algorithm}[H]
\caption{Probabilistic NaN Propagation \label{alg3}}
\begin{algorithmic}
\FOR{each feature $k$}
    \IF{\text{Bernoulli}$(p_k^{(y_i)})=1$}
        \STATE $\tilde{x}_k \leftarrow \text{NaN}$
    \ELSE
        \STATE interpolate if possible
    \ENDIF
\ENDFOR
\end{algorithmic}
\end{algorithm}

\section{Implementation}

OverNaN is designed as a lightweight extension to existing oversampling workflows rather than as a standalone computational framework. Its implementation prioritizes compatibility, transparency, and predictable resource usage over aggressive optimization. Nevertheless, several computational aspects are relevant in practice, particularly for moderate to large tabular datasets.

\subsection{Data Structures and Memory Use}

OverNaN operates on dense or masked tabular data represented as NumPy arrays or pandas dataframes. Missing values are represented explicitly as NaNs and are never implicitly converted or compressed. This choice preserves semantic clarity, but it implies that memory usage scales with the full dimensionality of the dataset, even when sparsity is high.

Synthetic samples are generated in batches and appended incrementally, rather than materializing all intermediate candidates at once. This limits peak memory consumption and avoids unnecessary duplication of large arrays. In particular, distance calculations and neighbour selection are performed on views or index sets wherever possible, rather than on copied arrays.

When class imbalance is extreme, the number of synthetic samples may exceed the size of the original dataset by an order of magnitude. In such cases, memory usage is dominated by the final resampled dataset rather than by the oversampling procedure itself. OverNaN does not implement streaming or out-of-core execution, and assumes that the resampled dataset fits in memory. This is consistent with its target use cases in small to medium-scale scientific applications.

\subsection{Neighbour Search and Computational Complexity}

For neighbour-based methods such as SMOTE and ADASYN, the dominant cost arises from nearest-neighbour queries. OverNaN delegates neighbour search to existing implementations, typically based on brute-force or tree-based methods from standard machine learning libraries. Distances are computed only on shared observed feature dimensions, which reduces arithmetic operations compared to full-dimensional distance calculations but introduces variable per-pair cost depending on missingness patterns.

In the worst case, neighbour search scales as $\mathcal{O}(N d)$ per query, where $N$ is the number of samples and $d$ the number of features. In practice, effective dimensionality is often reduced by missingness, and neighbour search dominates runtime only when both $N$ and $d$ are large. OverNaN does not modify the asymptotic complexity of the underlying oversampling algorithms.

\subsection{Pandas Compatibility}

OverNaN accepts both NumPy arrays and pandas containers. When a \texttt{DataFrame} and \texttt{Series} are passed in, the resampled outputs preserve the original container types, column names, and series name across all three methods. This supports integration with pipelines that depend on labeled features, such as feature attribution and downstream visualization tools.

\subsection{Parallel Execution}

OverNaN supports parallel execution at the level of synthetic sample generation. Independent interpolation operations are embarrassingly parallel, as each synthetic instance can be generated without reference to others once neighbours are selected. Where supported by the underlying oversampling backend, sample generation may be distributed across multiple CPU cores using thread-based or process-based parallelism.

Parallelism is applied conservatively. OverNaN avoids fine-grained locking or shared mutable state, and instead parallelizes over batches of synthetic samples. This reduces synchronization overhead and avoids race conditions related to shared data structures containing NaNs.
Neighbour search itself is not parallelized internally by OverNaN, but benefits indirectly from parallel implementations provided by external libraries when available. GPU acceleration is not supported, as the computational patterns involved are memory-bound and irregular, and typical problem sizes do not justify the overhead.

Sequential and parallel execution are compared on a moderately sized resampling task (1000 samples, 10 features, Table~\ref{tab:parallel}). The parallel implementation produces resampled arrays identical in shape and content to the sequential implementation. Wall-clock time depends on the per-task computational cost: ADASYNNaN, which performs adaptive density estimation, benefits substantially from parallelism (1.424\,s reduced to 0.331\,s), while SMOTENaN and ROSENaN are already fast in sequential form, so the per-batch parallel overhead can dominate at this problem size. Parallelism is therefore most useful when the underlying oversampling work per sample is non-trivial.

\begin{table}[h]
\centering
\caption{Sequential and parallel runtimes for resampling 1000 samples with 10 features. Output shapes match between modes for all methods.}
\label{tab:parallel}
\begin{tabular}{lccc}
\toprule
Method & Sequential (s) & Parallel (s) & Output shape \\
\midrule
SMOTENaN   & 0.272 & 2.627 & $(1000,10)$ \\
ADASYNNaN  & 1.424 & 0.331 & $(996,10)$  \\
ROSENaN    & 0.050 & 0.065 & $(1000,10)$ \\
\bottomrule
\end{tabular}
\end{table}

\subsection{Kernel-Based Sampling Costs}

For ROSE-style kernel-based oversampling, computational cost is dominated by random sampling and feature-wise perturbation rather than neighbour search. Memory access patterns are straightforward, and runtime scales linearly with the number of synthetic samples generated.

Bandwidth selection and covariance structure are treated as fixed hyperparameters. OverNaN does not estimate kernel densities explicitly, nor does it attempt to optimize bandwidth parameters at runtime. This design choice avoids additional memory overhead and numerical instability in the presence of missing values.

\subsection{Numerical Stability and NaN Propagation}

All arithmetic operations are performed in standard floating-point precision. NaNs are propagated deliberately according to defined rules, rather than filtered out or masked at the numerical level. OverNaN relies on the IEEE 754 standard behavior of NaNs, and does not introduce special sentinel values or bit-level encodings.

This design ensures numerical stability in the sense that no undefined arithmetic operations are performed. However, it places responsibility on downstream learners to handle NaNs correctly. OverNaN assumes compatibility with NaN-tolerant algorithms, such as tree-based ensemble methods, and does not attempt to enforce or emulate such behavior when it is absent.

\subsection{Reproducibility and Determinism}

Randomness in OverNaN arises from neighbour selection, interpolation coefficients, and kernel perturbations. All sources of randomness are controlled through explicit random seeds when provided. Given identical inputs, parameters, and random seeds, OverNaN produces deterministic outputs.

Parallel execution does not alter determinism, as synthetic samples are generated independently and aggregated using fixed ordering. This property is important for reproducibility in scientific workflows and aligns with the expectations of applied research domains.

\section{Testing and Validation}

OverNaN is distributed with a structured test suite covering basic functionality, sampling strategy options, missingness handling modes, kernel bandwidth behaviour, container compatibility, parallel execution, classifier integration, edge cases, and reproducibility. All tests are deterministic under a fixed random seed and exercise each of the three NaN-aware oversamplers: SMOTENaN, ADASYNNaN, and ROSENaN. Numerical values reported below are produced by the included test scripts.

\subsection{Basic Functionality}

A small imbalanced dataset of 120 samples and 10 features, with class distribution \{0: 100, 1: 20\} and an overall NaN rate of 20.17\%, is used to confirm that each method balances the classes and produces resampled arrays of the expected shape (Table~\ref{tab:basic}). All three methods rebalance the minority class to 100 samples without imputation. The resampled NaN rate increases for SMOTENaN and ADASYNNaN under the default \texttt{preserve\_pattern} strategy, because synthetic samples inherit missingness from either parent, while ROSENaN remains close to the original NaN rate due to its kernel-based perturbation around individual seeds.

\begin{table}[h]
\centering
\caption{Basic functionality on a small imbalanced dataset (120 samples, 10 features, 20.17\% NaN).}
\label{tab:basic}
\begin{tabular}{lccc}
\toprule
Method & Resampled shape & Class distribution & Resampled NaN \\
\midrule
SMOTENaN   & $(200,10)$ & \{0: 100, 1: 100\} & 27.80\% \\
ADASYNNaN  & $(200,10)$ & \{0: 100, 1: 100\} & 32.45\% \\
ROSENaN    & $(200,10)$ & \{0: 100, 1: 100\} & 20.35\% \\
\bottomrule
\end{tabular}
\end{table}

\subsection{Sampling Strategies}

All three oversamplers accept the same set of sampling-strategy specifications: the strings \texttt{auto}, \texttt{minority}, and \texttt{not majority}, a numeric ratio (e.g.\ 0.5), or an explicit dictionary of per-class target counts. Each specification produces the expected per-class output across SMOTENaN, ADASYNNaN, and ROSENaN. With a 100/20 starting distribution, \texttt{auto}, \texttt{minority}, and \texttt{not majority} all yield \{0: 100, 1: 100\}; ratio 0.5 yields \{0: 100, 1: 50\}; and the dictionary \{0: 100, 1: 80\} yields the requested counts. Behaviour is consistent across methods, confirming that the sampling-strategy interface is implemented uniformly.

\subsection{Missingness Handling}

The three NaN handling strategies described in Section~3 are tested on a separate dataset with an original NaN rate of 29.92\% (Table~\ref{tab:nan}). The \texttt{preserve\_pattern} strategy expands missingness in the resampled set, since synthetic samples inherit NaNs from either parent. The \texttt{interpolate} strategy reduces overall missingness by retaining observed values whenever at least one parent provides them. The \texttt{random\_pattern} strategy reproduces empirical per-feature missingness rates within each class. ROSENaN consistently reports lower resampled NaN rates than SMOTENaN or ADASYNNaN because it perturbs single seeds rather than interpolating between two parents.

\begin{table}[h]
\centering
\caption{Resampled NaN rate (\%) under the three missingness handling strategies. Original NaN rate is 29.92\%.}
\label{tab:nan}
\begin{tabular}{lccc}
\toprule
Strategy & SMOTENaN & ADASYNNaN & ROSENaN \\
\midrule
\texttt{preserve\_pattern}  & 41.75 & 41.36 & 32.30 \\
\texttt{interpolate}        & 22.35 & 24.09 & 17.95 \\
\texttt{random\_pattern}    & 31.20 & 33.79 & 22.60 \\
\bottomrule
\end{tabular}
\end{table}

\subsection{ROSE Shrinkage Behaviour}

The shrinkage parameter of ROSENaN controls the bandwidth of the Gaussian kernel used to perturb seed samples. Table~\ref{tab:shrinkage} reports the standard deviation of the synthetic samples for shrinkage values of 0.0, 0.5, 1.0, 1.5, and 2.0, where the original minority-class mean standard deviation is 0.9555. As expected, shrinkage values below one produce synthetic samples tighter than the seeds, while values above one produce broader synthetic distributions. The shrinkage parameter therefore offers a continuous control over the balance between fidelity to the seeds and exploration of nearby feature space.

\begin{table}[h]
\centering
\caption{Effect of the ROSENaN shrinkage parameter on synthetic-sample dispersion.}
\label{tab:shrinkage}
\begin{tabular}{cc}
\toprule
Shrinkage & Synthetic standard deviation \\
\midrule
0.0 & 0.9300 \\
0.5 & 0.9967 \\
1.0 & 1.1875 \\
1.5 & 1.4538 \\
2.0 & 1.7615 \\
\bottomrule
\end{tabular}
\end{table}

\subsection{Classifier Integration}

To verify that NaN-aware oversampling supports downstream learning, an XGBoost classifier is trained on a stratified 70/30 split of an imbalanced dataset with class distribution \{0: 350, 1: 35\} in the training set and \{0: 150, 1: 15\} in the test set. The baseline classifier, trained without oversampling, achieves a minority-class recall of 0.800 and minority $F_1$ of 0.889 (Table~\ref{tab:classifier}). All three NaN-aware oversamplers improve minority-class metrics. SMOTENaN and ROSENaN produce perfect recall and $F_1$ on the minority class, while ADASYNNaN improves recall to 0.867 and $F_1$ to 0.929. The test set is unmodified in all conditions; XGBoost handles its NaNs natively.

\begin{table}[h]
\centering
\caption{XGBoost test-set performance on an imbalanced classification task with NaNs, before and after NaN-aware oversampling. Test set has 150 majority and 15 minority samples.}
\label{tab:classifier}
\begin{tabular}{lcccc}
\toprule
Method & Minority recall & Minority $F_1$ & Weighted $F_1$ & Accuracy \\
\midrule
Baseline   & 0.800 & 0.889 & 0.981 & 0.98 \\
SMOTENaN   & 1.000 & 1.000 & 1.000 & 1.00 \\
ADASYNNaN  & 0.867 & 0.929 & 0.987 & 0.99 \\
ROSENaN    & 1.000 & 1.000 & 1.000 & 1.00 \\
\bottomrule
\end{tabular}
\end{table}

\subsection{Comparative Behaviour on a Controlled Synthetic Dataset}

\begin{figure}[h]
\centering
\includegraphics[width=\textwidth]{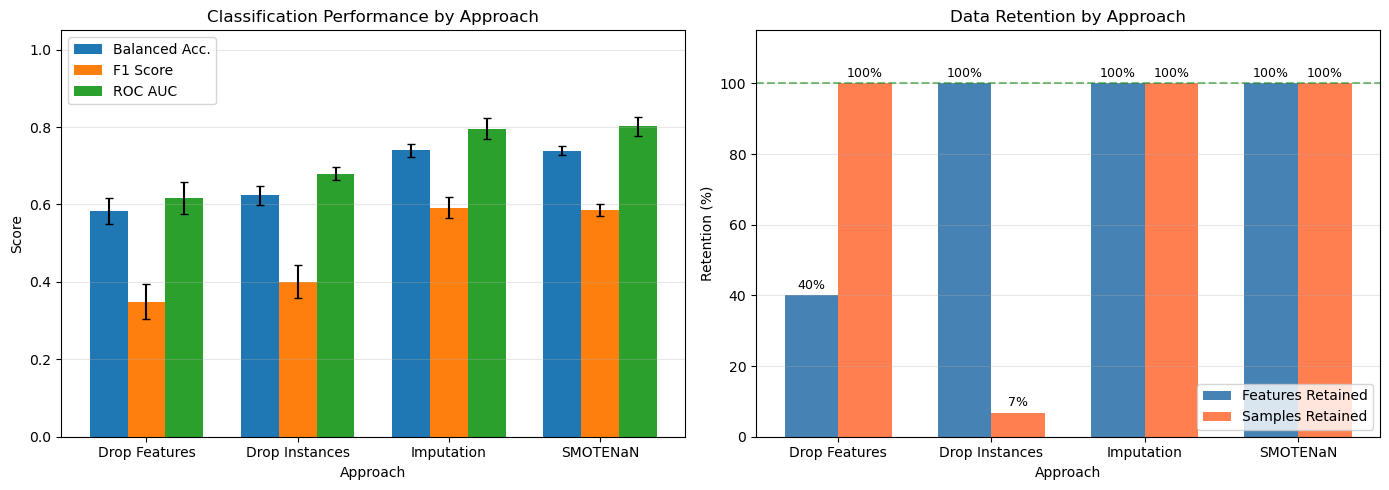}
\caption{Classification performance and data retention by NaN handling strategy. Dropping features retains only 40\% of features; dropping instances retains only 7\% of samples; imputation and SMOTENaN both retain all features and samples.}
\label{fig:synthetic}
\end{figure}

A controlled synthetic comparison contrasts SMOTENaN against three common alternatives: removing features that contain any NaNs, removing samples that contain any NaNs, and mean imputation prior to standard SMOTE. The dataset comprises 2000 samples and 20 features, of which 12 carry concentrated missingness, with an overall NaN rate of 11.9\% and an imbalance ratio of approximately 4:1. Evaluation uses 5-fold stratified cross-validation with XGBoost as the downstream learner. Results are summarized in Table~\ref{tab:synthetic} and Figure~\ref{fig:synthetic}.

\begin{table}[h]
\centering
\caption{Synthetic comparison of NaN handling strategies under 5-fold stratified cross-validation with XGBoost.}
\label{tab:synthetic}
\begin{tabular}{lccccc}
\toprule
Approach              & Bal.\ Acc.\         & $F_1$               & Precision           & Recall              & ROC AUC \\
\midrule
Drop Features + SMOTE & $0.583\pm0.033$ & $0.349\pm0.045$ & $0.298\pm0.031$ & $0.422\pm0.068$ & $0.617\pm0.041$ \\
Drop Instances + SMOTE& $0.623\pm0.024$ & $0.401\pm0.042$ & $0.414\pm0.037$ & $0.391\pm0.053$ & $0.680\pm0.017$ \\
Imputation + SMOTE    & $0.740\pm0.016$ & $0.592\pm0.027$ & $0.613\pm0.044$ & $0.575\pm0.033$ & $0.795\pm0.028$ \\
SMOTENaN              & $0.739\pm0.011$ & $0.586\pm0.015$ & $0.590\pm0.026$ & $0.585\pm0.029$ & $0.802\pm0.024$ \\
\bottomrule
\end{tabular}
\end{table}

Dropping features retains only 8 of 20 columns and produces the weakest classifier. Dropping instances retains all 20 features but only about 7\% of samples (137 of 2000), restricting the training pool. Imputation and SMOTENaN reach comparable balanced accuracy and $F_1$, but SMOTENaN obtains the highest ROC AUC (0.802) and the smallest cross-fold variance, while preserving rather than fabricating values for missing entries.

\subsection{Edge Cases}

The test suite verifies graceful handling of degenerate inputs. With only one minority sample per class, all three oversamplers complete and return at least the requested distribution. When an entire seed has all-NaN features, SMOTENaN and ADASYNNaN return shape $(4,3)$ and ROSENaN returns shape $(6,3)$, both without raising exceptions. With a deliberately high NaN rate of 50\%, the resampled NaN rates remain controlled (60.00\%, 60.29\%, and 47.80\% for SMOTENaN, ADASYNNaN, and ROSENaN respectively). With a single feature, all three methods complete and return the expected shapes.

\subsection{Reproducibility}

Reproducibility is verified by re-running each method twice with the same random seed and confirming that both shape and element-wise values match between runs. SMOTENaN, ADASYNNaN, and ROSENaN all pass this test. Combined with the deterministic neighbour-search and interpolation procedure, this means that a single random seed fully determines the resampled output, supporting independent verification and consistent reporting.

\section{Benchmarking on OpenML}

To assess OverNaN on real datasets with naturally occurring missing values, we benchmark SMOTENaN, ADASYNNaN, and ROSENaN against an unmodified XGBoost baseline on five publicly available OpenML datasets~\cite{vanschoren2014openml}, each containing class imbalance and missingness that arises from the data-collection process rather than artificial corruption. Datasets and their characteristics are summarised in Table~\ref{tab:bench-datasets}. Evaluation uses 5-fold stratified cross-validation with a fixed random seed.

\begin{table}[h]
\centering
\caption{OpenML benchmark datasets~\cite{vanschoren2014openml,dua2019uci}. Imbalance ratio (IR) is reported as majority-to-minority.}
\label{tab:bench-datasets}
\begin{tabular}{lccccc}
\toprule
Dataset & OpenML ID & Samples & Features & NaN rate & IR \\
\midrule
labor          & 4     & 57   & 16  & 35.7\% & 1.85 \\
cylinder-bands & 6332  & 540  & 37  & 5.0\%  & 1.37 \\
Titanic        & 40945 & 1309 & 13  & 22.7\% & 1.62 \\
soybean        & 1023  & 683  & 35  & 9.8\%  & 6.42 \\
car-evaluation & 1017  & 452  & 279 & 0.3\%  & 1.18 \\
\bottomrule
\end{tabular}
\end{table}

\subsection{Per-Dataset Results}

Tables~\ref{tab:bench-labor}--\ref{tab:bench-car} report balanced accuracy, macro $F_1$, minority $F_1$, geometric mean (G-Mean), and ROC AUC for each method on each dataset, with mean and standard deviation across the five folds.

\begin{table}[h]
\centering
\caption{labor (NaN: 35.7\%, IR: 1.9).}
\label{tab:bench-labor}
\begin{tabular}{lccccc}
\toprule
Method & Bal.\ Acc.\ & $F_1$ Macro & $F_1$ Min & G-Mean & ROC AUC \\
\midrule
Baseline   & $0.911\pm0.077$ & $0.920\pm0.074$ & $0.893\pm0.096$ & $0.907\pm0.080$ & $0.959\pm0.056$ \\
SMOTENaN   & $0.896\pm0.065$ & $0.901\pm0.062$ & $0.871\pm0.078$ & $0.892\pm0.067$ & $0.965\pm0.055$ \\
ADASYNNaN  & $0.868\pm0.085$ & $0.866\pm0.094$ & $0.836\pm0.107$ & $0.862\pm0.085$ & $0.986\pm0.029$ \\
ROSENaN    & $0.911\pm0.077$ & $0.920\pm0.074$ & $0.893\pm0.096$ & $0.907\pm0.080$ & $0.971\pm0.057$ \\
\bottomrule
\end{tabular}
\end{table}

\begin{table}[h]
\centering
\caption{cylinder-bands (NaN: 5.0\%, IR: 1.4).}
\label{tab:bench-cylinder}
\begin{tabular}{lccccc}
\toprule
Method & Bal.\ Acc.\ & $F_1$ Macro & $F_1$ Min & G-Mean & ROC AUC \\
\midrule
Baseline   & $0.796\pm0.034$ & $0.799\pm0.034$ & $0.754\pm0.053$ & $0.787\pm0.045$ & $0.890\pm0.015$ \\
SMOTENaN   & $0.813\pm0.039$ & $0.817\pm0.040$ & $0.777\pm0.057$ & $0.806\pm0.048$ & $0.895\pm0.022$ \\
ADASYNNaN  & $0.820\pm0.034$ & $0.823\pm0.034$ & $0.786\pm0.047$ & $0.814\pm0.040$ & $0.902\pm0.015$ \\
ROSENaN    & $0.823\pm0.042$ & $0.827\pm0.043$ & $0.791\pm0.055$ & $0.818\pm0.047$ & $0.908\pm0.021$ \\
\bottomrule
\end{tabular}
\end{table}

\begin{table}[h]
\centering
\caption{Titanic (NaN: 22.7\%, IR: 1.6).}
\label{tab:bench-titanic}
\begin{tabular}{lccccc}
\toprule
Method & Bal.\ Acc.\ & $F_1$ Macro & $F_1$ Min & G-Mean & ROC AUC \\
\midrule
Baseline   & $0.965\pm0.011$ & $0.967\pm0.010$ & $0.959\pm0.012$ & $0.965\pm0.012$ & $0.993\pm0.003$ \\
SMOTENaN   & $0.965\pm0.008$ & $0.963\pm0.005$ & $0.954\pm0.006$ & $0.964\pm0.008$ & $0.994\pm0.003$ \\
ADASYNNaN  & $0.961\pm0.008$ & $0.960\pm0.005$ & $0.951\pm0.006$ & $0.961\pm0.008$ & $0.993\pm0.004$ \\
ROSENaN    & $0.967\pm0.015$ & $0.968\pm0.012$ & $0.960\pm0.016$ & $0.967\pm0.015$ & $0.995\pm0.003$ \\
\bottomrule
\end{tabular}
\end{table}

\begin{table}[h]
\centering
\caption{soybean (NaN: 9.8\%, IR: 6.4).}
\label{tab:bench-soybean}
\begin{tabular}{lccccc}
\toprule
Method & Bal.\ Acc.\ & $F_1$ Macro & $F_1$ Min & G-Mean & ROC AUC \\
\midrule
Baseline   & $0.930\pm0.038$ & $0.944\pm0.028$ & $0.903\pm0.050$ & $0.927\pm0.041$ & $0.993\pm0.005$ \\
SMOTENaN   & $0.948\pm0.047$ & $0.946\pm0.042$ & $0.907\pm0.050$ & $0.946\pm0.050$ & $0.994\pm0.005$ \\
ADASYNNaN  & $0.940\pm0.037$ & $0.948\pm0.032$ & $0.910\pm0.057$ & $0.938\pm0.039$ & $0.993\pm0.006$ \\
ROSENaN    & $0.950\pm0.040$ & $0.952\pm0.036$ & $0.917\pm0.063$ & $0.948\pm0.042$ & $0.994\pm0.004$ \\
\bottomrule
\end{tabular}
\end{table}

\begin{table}[h]
\centering
\caption{car-evaluation (NaN: 0.3\%, IR: 1.2).}
\label{tab:bench-car}
\begin{tabular}{lccccc}
\toprule
Method & Bal.\ Acc.\ & $F_1$ Macro & $F_1$ Min & G-Mean & ROC AUC \\
\midrule
Baseline   & $0.817\pm0.060$ & $0.817\pm0.059$ & $0.794\pm0.073$ & $0.813\pm0.062$ & $0.881\pm0.072$ \\
SMOTENaN   & $0.808\pm0.062$ & $0.808\pm0.062$ & $0.784\pm0.078$ & $0.803\pm0.065$ & $0.879\pm0.065$ \\
ADASYNNaN  & $0.817\pm0.060$ & $0.817\pm0.059$ & $0.794\pm0.073$ & $0.813\pm0.062$ & $0.881\pm0.072$ \\
ROSENaN    & $0.832\pm0.063$ & $0.833\pm0.063$ & $0.812\pm0.076$ & $0.829\pm0.065$ & $0.898\pm0.051$ \\
\bottomrule
\end{tabular}
\end{table}

\subsection{Average Performance and Runtime}

Aggregating across the five datasets (Table~\ref{tab:bench-summary}), ROSENaN achieves the highest mean balanced accuracy, macro $F_1$, and ROC AUC, while remaining within a small constant factor of the baseline runtime. SMOTENaN provides modest gains over baseline. ADASYNNaN matches or slightly underperforms baseline on average for these datasets but consistently improves ROC AUC, suggesting better-calibrated probability estimates at the cost of harder thresholded decisions. The improvement over baseline in percentage points (Table~\ref{tab:bench-improve}) shows ROSENaN as the most consistently beneficial method across the four imbalance-aware metrics.

\begin{table}[h]
\centering
\caption{Average performance and runtime across the five OpenML benchmark datasets.}
\label{tab:bench-summary}
\begin{tabular}{lcccccc}
\toprule
Method & Bal.\ Acc.\ & $F_1$ Macro & $F_1$ Min & G-Mean & ROC AUC & Time (s) \\
\midrule
Baseline   & 0.8836 & 0.8893 & 0.8604 & 0.8798 & 0.9432 & 0.3  \\
SMOTENaN   & 0.8860 & 0.8869 & 0.8584 & 0.8822 & 0.9455 & 3.5  \\
ADASYNNaN  & 0.8811 & 0.8830 & 0.8555 & 0.8775 & 0.9509 & 16.5 \\
ROSENaN    & 0.8967 & 0.8999 & 0.8746 & 0.8937 & 0.9533 & 0.7  \\
\bottomrule
\end{tabular}
\end{table}

\begin{table}[h]
\centering
\caption{Improvement over baseline, in percentage points, averaged across the five datasets.}
\label{tab:bench-improve}
\begin{tabular}{lcccc}
\toprule
Method & $\Delta$ Bal.\ Acc.\ & $\Delta F_1$ Macro & $\Delta F_1$ Min & $\Delta$ G-Mean \\
\midrule
SMOTENaN   & $+0.24$ & $-0.24$ & $-0.20$ & $+0.24$ \\
ADASYNNaN  & $-0.25$ & $-0.63$ & $-0.49$ & $-0.23$ \\
ROSENaN    & $+1.31$ & $+1.06$ & $+1.42$ & $+1.39$ \\
\bottomrule
\end{tabular}
\end{table}

\subsection{OverNaN versus Impute-then-Oversample}

A direct comparison between NaN-aware oversampling and the conventional impute-then-oversample pipeline is performed on the OpenML \textit{anneal} dataset~\cite{vanschoren2014openml,dua2019uci} (898 samples, 38 features, imbalance ratio 3.20, NaN rate 64.98\%). Mean imputation precedes standard SMOTE, ADASYN, or ROSE in the impute-then-oversample variants; OverNaN performs equivalent oversampling without imputation. Results in Table~\ref{tab:anneal} show that the two pipelines produce statistically indistinguishable balanced accuracy and macro $F_1$ on this dataset, with all variants well within one standard deviation of each other. Imputation is therefore not a precondition for competitive performance, and avoiding imputation removes the need to introduce values that were not measured.

\begin{table}[h]
\centering
\caption{Impute-then-oversample versus OverNaN on the OpenML \textit{anneal} dataset (NaN rate 64.98\%).}
\label{tab:anneal}
\begin{tabular}{lcc}
\toprule
Approach          & Bal.\ Acc.\         & $F_1$ Macro         \\
\midrule
Baseline          & $0.985\pm0.005$ & $0.986\pm0.003$ \\
Impute + SMOTE    & $0.991\pm0.007$ & $0.988\pm0.009$ \\
Impute + ADASYN   & $0.989\pm0.004$ & $0.986\pm0.006$ \\
Impute + ROSE     & $0.985\pm0.012$ & $0.986\pm0.009$ \\
OverNaN-SMOTE     & $0.989\pm0.008$ & $0.986\pm0.007$ \\
OverNaN-ADASYN    & $0.990\pm0.009$ & $0.989\pm0.008$ \\
OverNaN-ROSE      & $0.986\pm0.007$ & $0.983\pm0.006$ \\
\bottomrule
\end{tabular}
\end{table}

\section{Application: Graphene Oxide Nanoflakes}

The graphene oxide example uses bond and ring statistics extracted from a subset of the Neutral Graphene Oxide Data Set~\cite{barnard2019grapheneoxide} to predict the presence or absence of carboxyl (COOH) functional groups on hexagonal nanoflakes. Bond and ring statistics describe local atomic connectivity and topology and were defined for this family of structures in~\cite{motevalli2019representative}. The subset is restricted to large hexagonal nanoflakes (C\,$=$\,1014 carbon atoms), giving 776 structures. The 232 features comprise 192 bond-statistics features and 40 ring-statistics features.

\subsection{Why Carboxyl Groups, and Why Missingness is Meaningful}

Carboxyl groups are chemically significant functional groups that influence the hydrophilicity, dispersibility, and reactivity of graphene oxide. They are also relatively rare in this subset: 200 structures lack carboxyl groups while 576 contain at least one, giving an imbalance ratio of approximately 2.9:1 in favour of the carboxyl-bearing class. Predicting carboxyl presence requires resolving subtle differences in local structural patterns.

Missing values arise structurally rather than by experimental failure. Specific bond types and ring configurations may simply be absent in particular structures (for example, an O-O bond length is undefined in a nanoflake without any O-O bonds). The dataset accordingly contains a substantial 39.8\% NaN rate, with 123 of the 232 features (53\%) carrying missingness, and every one of the 776 samples containing at least one NaN. This makes carboxyl prediction an informative test case for NaN-aware oversampling: imputing absent bonds with mean or median values would inject artificial structural information that does not exist in any of the underlying nanoflakes.

\subsection{Comparison with Drop Features and Imputation}

Three pipelines are compared under 5-fold stratified cross-validation with XGBoost (\texttt{n\_estimators}=100, \texttt{max\_depth}=4, \texttt{learning\_rate}=0.1):
\begin{enumerate}
\item Drop Features + SMOTE: features containing any NaN are removed, then standard SMOTE is applied to the remaining features.
\item Imputation + SMOTE: NaN entries are filled by median imputation, then standard SMOTE is applied.
\item SMOTENaN: NaN-aware SMOTE under the \texttt{preserve\_pattern} strategy, applied directly to the data.
\end{enumerate}

\begin{figure}[h]
\centering
\includegraphics[width=\textwidth]{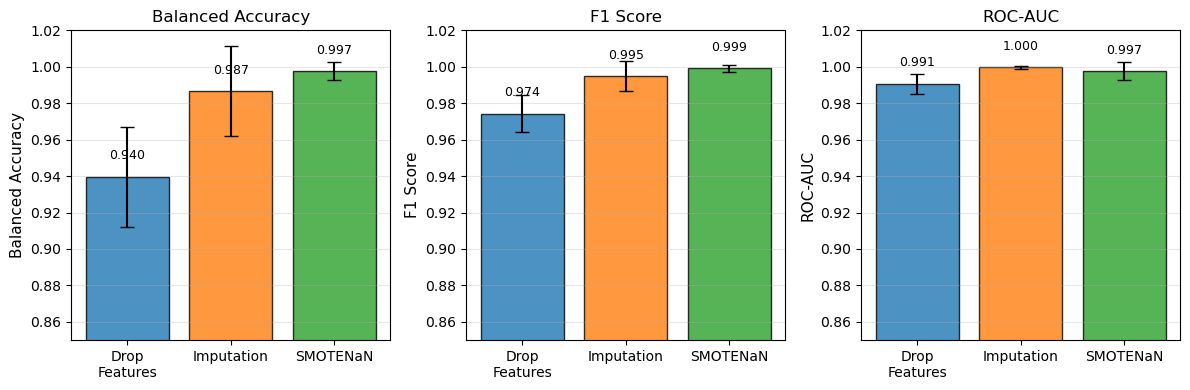}
\caption{Carboxyl prediction on hexagonal graphene oxide nanoflakes. SMOTENaN reaches the highest balanced accuracy and $F_1$ score with the smallest cross-fold variance, while preserving the structural meaning of missing entries.}
\label{fig:graphene}
\end{figure}

Data retention is summarised in Table~\ref{tab:go-retention} and classification metrics in Table~\ref{tab:go-results} and Figure~\ref{fig:graphene}. Dropping features removes 53\% of the feature space, retaining only 109 of the 232 bond and ring statistics, which corresponds to discarding most of the chemically informative columns. Imputation retains all features and samples but replaces 39.8\% of all entries with synthetic median values. SMOTENaN retains all features and all samples without introducing values that were not observed.

\begin{table}[h]
\centering
\caption{Data retention by pipeline on the graphene oxide carboxyl-prediction task.}
\label{tab:go-retention}
\begin{tabular}{lccl}
\toprule
Approach        & Features retained & Samples retained & Information loss \\
\midrule
Drop Features   & 109/232 (47\%)    & 776/776 (100\%)  & Removes 53\% of features \\
Imputation      & 232/232 (100\%)   & 776/776 (100\%)  & Replaces 39.8\% of entries \\
SMOTENaN        & 232/232 (100\%)   & 776/776 (100\%)  & None (NaNs preserved) \\
\bottomrule
\end{tabular}
\end{table}

\begin{table}[h]
\centering
\caption{Carboxyl prediction on hexagonal graphene oxide nanoflakes (5-fold stratified CV, XGBoost).}
\label{tab:go-results}
\begin{tabular}{lccc}
\toprule
Approach            & Balanced Accuracy   & $F_1$ Score         & ROC AUC             \\
\midrule
Drop Features + SMOTE & $0.940\pm0.027$ & $0.974\pm0.010$ & $0.991\pm0.006$ \\
Imputation + SMOTE    & $0.987\pm0.025$ & $0.995\pm0.008$ & $1.000\pm0.001$ \\
SMOTENaN              & $0.997\pm0.005$ & $0.999\pm0.002$ & $0.997\pm0.005$ \\
\bottomrule
\end{tabular}
\end{table}

SMOTENaN attains the highest balanced accuracy ($0.997 \pm 0.005$) and $F_1$ score ($0.999 \pm 0.002$), with a markedly smaller cross-fold variance than both alternatives. Drop Features carries the largest variance and the lowest mean, consistent with the loss of more than half the feature space. Imputation matches SMOTENaN in mean ROC AUC but with substantially larger variance, reflecting the fold-to-fold sensitivity of median imputation under high NaN rates. The gain in mean accuracy from SMOTENaN over imputation is modest in absolute terms but is achieved without committing to any specific value for an unobserved entry, which is desirable when the intent is to characterize structural chemistry rather than smooth over it.

\subsection{Discussion}

The graphene oxide example illustrates a setting in which missingness is not noise but signal. Whether a structure contains an O-O bond is a property of the structure, not of the measurement. Imputing the corresponding bond length to the dataset mean obscures this distinction, while dropping the feature discards an informative descriptor. SMOTENaN supports the alternative position that bond and ring statistics may legitimately be absent, and propagates that absence into synthetic samples without introducing values that have no chemical referent.

%\section{Scope and Limitations}
While the graphene oxide dataset  is not considered to be small in computational nanotechnology, OverNaN is not optimized for very large datasets, nor is it intended for deployment in high-throughput or real-time systems. Its implementation reflects the priorities of exploratory and model-development workflows, where interpretability, control over assumptions, and reproducibility outweigh raw throughput.  The absence of aggressive memory compression, GPU acceleration, or distributed execution is a deliberate design choice, consistent with the goal of preserving missingness structure and maintaining compatibility with established machine learning toolchains.

It is also prudent to point out that OverNaN is not a universal solution to missing data. When missing values must be estimated for downstream tasks, imputation may still be necessary. OverNaN does not model missingness mechanisms, nor does it infer unobserved values. The framework is most appropriate for small, imbalanced datasets where missingness is systematic and informative, and where downstream learners can tolerate NaNs. Its effectiveness depends on thoughtful selection of missingness-handling strategies aligned with domain knowledge.

\section{Conclusion}

Missingness is often treated as an inconvenience to be eliminated before modelling. In many applied settings, this assumption is unwarranted. OverNaN provides a simple, explicit mechanism for addressing class imbalance without erasing missingness structure, offering an alternative to deletion and imputation in domains where what is missing may matter as much as what is observed. By making missingness a first-class consideration in oversampling, OverNaN supports more faithful representations of real, incomplete data, and encourages practitioners to treat uncertainty explicitly rather than conceal it.

\section*{Availability}

OverNaN is open-source and available at: \url{https://github.com/amaxiom/OverNaN}, or via \texttt{pip install overnan}.

\clearpage
\bibliographystyle{unsrt}
\bibliography{references.bib}

\end{document}